\documentclass[conference]{IEEEtran}
\IEEEoverridecommandlockouts
\usepackage{cite}
\usepackage{amsmath,amssymb,amsfonts}
\usepackage{algorithmic}
\usepackage{graphicx}
\usepackage{textcomp}
\usepackage{xcolor}
\def\BibTeX{{\rm B\kern-.05em{\sc i\kern-.025em b}\kern-.08em
    T\kern-.1667em\lower.7ex\hbox{E}\kern-.125emX}}

\newcommand{\citep}[1]{(\cite{#1})}
\newcommand{\citet}[1]{\cite{#1}}

\usepackage{amsmath}
\usepackage{graphicx}
\usepackage{algorithm}
\usepackage{algorithmic}
\usepackage{color}
\usepackage{booktabs}

\DeclareMathOperator{\mimi}{minimize\ \ \ \ }
\DeclareMathOperator{\sut}{subject\ to\ \ \ \ }
\DeclareMathOperator{\eps}{eps}
\DeclareMathOperator{\lineps}{lineps}

\newcommand\blfootnote[1]{%
  \begingroup
  \renewcommand\thefootnote{}\footnote{#1}%
  \addtocounter{footnote}{-1}%
  \endgroup
}

\begin{document}

\title{Local Propagation in \\ Constraint-based Neural Networks}

\author{

\IEEEauthorblockN{Giuseppe Marra\IEEEauthorrefmark{1}, Matteo Tiezzi\IEEEauthorrefmark{2}, Stefano Melacci\IEEEauthorrefmark{2}, Alessandro Betti\IEEEauthorrefmark{2}, Marco Maggini\IEEEauthorrefmark{2}, Marco Gori\IEEEauthorrefmark{2}}\\
\IEEEauthorblockA{\IEEEauthorrefmark{1}\textit{Dept. of Information Engineering} \\
\textit{University of Florence}\\
Florence, Italy \\
\small{\texttt{\{g.marra,alessandro.betti\}@unifi.it}}}
\\
\IEEEauthorblockA{\IEEEauthorrefmark{2}\textit{Dept. of Information Engineering and Mathematical Sciences} \\
\textit{University of Siena}\\
Siena, Italy \\
\small{\texttt{\{mtiezzi,mela,maggini,marco\}@diism.unisi.it}}}
}

\maketitle

\begin{abstract}
In this paper we study a constraint-based representation of neural network architectures. 
We cast the learning problem in the Lagrangian framework and we investigate a simple optimization procedure
that is well suited to fulfil the so-called architectural constraints, learning from the available supervisions.
The computational structure of the proposed Local Propagation (LP) algorithm is based on the search for saddle points in
the adjoint space composed of weights, neural outputs, and Lagrange multipliers. 
All the updates of the model variables are locally performed, so that LP is fully parallelizable over the neural units, circumventing
the classic problem of gradient vanishing in deep networks. 
The implementation of popular neural models is described in the context of LP, together with those conditions that trace a natural connection with Backpropagation. 
We also investigate the setting in which we tolerate bounded violations of the architectural constraints, and 
we provide experimental evidence that LP is a 
feasible approach to train shallow and deep networks, opening the road to further investigations
on more complex architectures, easily describable by constraints.
\end{abstract}

\section{Introduction}
\label{sec:intro}
\blfootnote{Accepted for publication at the IEEE International Joint Conference on Neural Networks (IJCNN) 2020 (DOI: TBA).

© 2020 IEEE. Personal use of this material is permitted. Permission from IEEE must be obtained for all other uses, in any current or future media, including reprinting/republishing this material for advertising or promotional purposes, creating new collective works, for resale or redistribution to servers or lists, or reuse of any copyrighted component of this work in other works.}
In the last years, neural networks have become extremely widespread models, due to their role in several important achievements of the Machine Learning community \cite{sm}. If we consider the recent scientific contributions in the field, it is often the case of new neural architectures that are designed to solve the task at hand \cite{Yang_2016_CVPR,xu2016ask}, or of new architectures that are created as alternatives to existing models \cite{vaswani2017attention}.
Backpropagation \cite{rumelhart1988learning} is assumed to be the ``de facto'' algorithm for training neural nets.

In this paper, we are inspired by the ideas of describing learning using the unifying notion of ``constraint'' \cite{gori2017machine,gnecco2015foundations}, and we nicely intercept the work of \cite{lecun1988theoretical}, where a theoretical framework for Backpropagation is studied in a Lagrangian formulation of learning. In particular, we regard the neural architecture as a set of constraints that correspond with the neural equations, which enforce the consistency between the input and the output variables by means of the corresponding weights of the synaptic connections. 
However, differently from \cite{lecun1988theoretical}, we do not only focus on the derivation of Backpropation in the Lagrangian framework, and we introduce a novel approach to learning that explores the search in the adjoint space that is characterized by the triple $(w, x, \lambda)$, i.e., (weights, neuron outputs, and Lagrangian multipliers). The idea of training networks represented with constraints and extending the space of learnable parameters has been originally introduced by \cite{carreira2014distributed}, using an optimization scheme based on quadratic penalty. The approach of \cite{carreira2014distributed} is built on the idea of finding an inexact solution of the original learning problem, and it relies on a post-processing procedure that refines the last-layer connections. The related approach of \cite{taylor2016training} involves closed-form solutions, but most of the architectural constraints are softly enforced, and further additional variables are introduced to parametrize the neuron activations. Other approaches followed these seminal works to implement constraining schemes for block-wise optimization of neural networks \cite{noia}.

Differently, in this paper we propose a hard-constraining scheme based on the augmented Lagrangian and on the optimization procedure of \cite{platt}, in which we search for saddle points in the adjoint space by a differential optimization process. This procedure is very easy to implement. 

The obtained results show that constraint-based networks, even if optimized with the proposed simple strategy, can be trained in an effective way. However, the main goal of the paper is not to show improved performance w.r.t. Backpropagation, with which it shares the same Lagrangian derivation, but to propose an optimization scheme for the weights of a neural network which shows new and promising properties. Indeed, it turns out that the gradient descent w.r.t  to the variables $(w, x)$ and the gradient ascent w.r.t to the multipliers $\lambda$ give rise to a truly \textit{local} algorithm for the variable updates that we refer to as Local Propagation (LP). By avoiding long dependencies among variable gradients, this method nicely circumvents the vanishing gradient problem in optimizing neural networks. Moreover, the local nature of the proposed algorithm enables  the parallelization of the training computations over the neural units. Finally, by interpreting Lagrange multipliers as the reaction to single neural computations, the proposed scheme opens the door to new methods for architecture search in the deep learning scenario.

Differently from \cite{carreira2014distributed,taylor2016training}, we study the connections with BackPropagation, performing an extended comparison on several benchmarks.
Moreover, we do not optimize variables associated to the activation score of each neuron. We use the Lagrangian approach to find a solution of the optimization problem that includes hard constraints. Instead of following a soft optmization procedure, we relax the constraints with $\epsilon$-based functions, where the tolerance is fixed and defined in advance. 


This paper makes three important contributions. First, it introduces a local algorithm (LP) for training neural networks described by means of the so-called architectural constraints, evaluating a simple optimization approach. 
Second, the implementation of popular neural models is described in the context of LP, together with the conditions under which we can see the natural connection with Backpropagation. Third, we investigate the setting in which we tolerate bounded violations of the architectural constraints, and we provide experimental evidence that LP is a feasible approach to train shallow and deep networks. LP also opens the road to further investigations on more complex architectures, easily describable by means of constraints.

\section{Constraint-based Neural Networks}
\label{sec:constr}
We are given $N$ supervised pairs $(x_{0,i}, y_i)$, $i=1,\ldots,N$, and we consider a generic neural architecture described by a Directed Acyclic Graph (DAG) that, in the context of this paper and without any loss of generality, is a Multi Layer Perceptron (MLP) with $H$ hidden layers.
The output of a generic hidden layer $\ell \in [1, H]$ for the $i$-example is indicated with $x_{\ell,i}$ that is a column vector with a number of components equal to the number of hidden units in such layer. We also have that $x_{0,i}$ is the input signal, and $x_{\ell,i} = \sigma(W_{\ell-1} x_{\ell-1,i})$, where $\sigma(\cdot)$ is the activation function that is intended to operate element-wise on its vectorial argument (we assume that this property holds in all the following functions). The matrix $W_{\ell-1}$ collects the weights linking layer $\ell-1$ to $\ell$. We avoid introducing bias terms in the argument of $\sigma(\cdot)$, to simplify the notation. 
The function $V(W_{H}x_{H,i}, y_i)$ computes the loss on the $i$-th supervised pair, and, when summed up for all the $N$ pairs, it yields the objective function that is minimized by the learning algorithm. In the case of classic neural networks, the variables involved in the optimization are the weights $W_{\ell}$, $\forall \ell$. 

We formulate the learning problem by describing the network architecture with a set of constraints. In particular, all the $x_{\ell,i}$'s become variables of the learning problem, and they are constrained to fulfil the so-called (hard) \textit{architectural constraints} $x_{\ell,i} = \sigma(W_{\ell-1} x_{\ell-1,i})$, i.e., \footnote{Without any loss of generality, we could also introduce the same constraint in the output layer ($\ell = H + 1$).}
\begin{align}
\nonumber \mimi & \sum_{i=1}^{N} V(W_{H}x_{H,i}, y_i)\\
\sut & \mathcal{G}(x_{\ell,i} - \sigma(W_{\ell-1} x_{\ell-1,i})) = 0, \quad \forall (i,\ell)
\label{con}
\end{align}
being $\mathcal{G}(\cdot)$ a generic function such that $\mathcal{G}(0)=0$, and that is only used to differently weight the mismatch between $x_{\ell,i}$ and $\sigma(W_{\ell-1} x_{\ell-1,i})$.
We will also make use of the notation $\mathcal{G}_{\ell,i}$ to compactly indicate the left-hand side of Eq. (\ref{con}).
In the Lagrangian framework \cite{bertsekas2014constrained}, if  $\lambda_{\ell,i}$ are the Lagrange multipliers associated to each architectural constraint, then we can write Lagrangian function $\mathcal{L}$ as
\begin{equation}
\mathcal{L}(\mathcal{W}, \mathcal{X}, \Lambda) = \sum_{i=1}^{N} \Bigl( V(W_{H}x_{H,i}, y_i) + \sum_{\ell=1}^{H} \lambda^{T}_{\ell,i} \mathcal{G}_{\ell,i} \Bigr) \! ,
\label{lagrangian}
\end{equation}
where we only emphasized the dependance on the set of variables that are involved in the learning process: $\mathcal{W}$ is the set of all the network weights; $\mathcal{X}$ is the set of all the $x_{\ell,i}$'s variables; $\Lambda$ collects the Lagrange multipliers ($T$ is the transpose operator).
The Lagrangian $\mathcal{L}$ can also be augmented with a squared $L_{2}$ norm regularizer on the network weights, scaled by a positive factor $c$.

\section{Local Propagation}
\label{lp}
Despite the variety of popular approaches that can be used to solve the constrained problem above \cite{bertsekas2014constrained}, we decided to focus on the optimization procedure studied in the context of neural networks in \cite{platt}. In the proposed Local Propagation (LP) algorithm, learning consists in a ``differential optimization'' process that converges towards a saddle point of Eq. (\ref{lagrangian}), minimizing it with respect to $\mathcal{W}$ and $\mathcal{X}$, and maximizing it with respect to $\Lambda$. The whole procedure is very simple, and it consists in performing a gradient-descent step to update $\mathcal{W}$ and $\mathcal{X}$, and a gradient-ascent step to update $\Lambda$, until we converge to a stationary point. As it will become clear shortly, each iteration of the optimization algorithm is $\mathcal{O}(|\mathcal{W}|)$ (without considering the number of examples), that is, it exhibits the same optimal asymptotical property of Backpropagation.

We initialize the variables in $\mathcal{X}$ and $\Lambda$ to zero, while the weights $\mathcal{W}$ are randomly chosen. For this reason,  at the beginning of the optimization, the degree of fulfillment is the same for all the architectural constraints $\mathcal{G}_{\ell,i}$, $\ell > 1$, in every unit of all layers and for all examples, while only $V(\cdot,\cdot)$ and $\mathcal{G}_{1,i}$ (that is, the farthest portions of the architecture) contribute to the Lagrangian of Eq.~(\ref{lagrangian}).
In the case of Backpropagation, the outputs of the neural units are the outcome of the classic forward step, while in the training stage of LP the evolution of the variables in $\mathcal{X}$ is dictated by gradient-based optimization. Once LP has converged, the architectural constraints of Eq.~(\ref{con}) are fulfilled, so that we can easily devise the values in $\mathcal{X}$ with the same forward step of Backpropagation-trained networks. In other words, we can consider LP as an algorithm to train the network weights while still relying on the classic forward pass during inference.

\begin{figure*}
\centering
\includegraphics[width=0.42\textwidth]{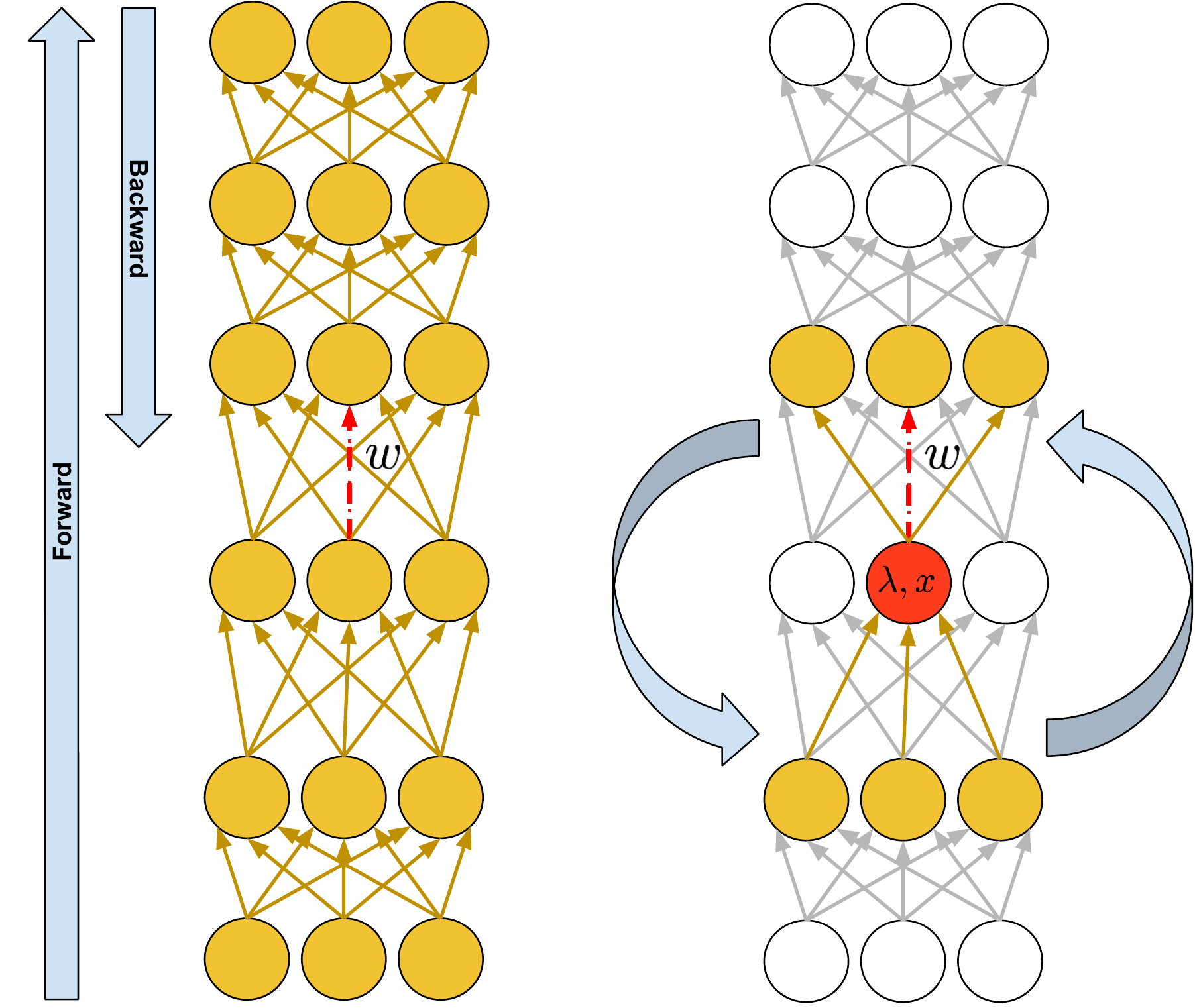} \hskip 0.5cm {\unskip\ \vrule\ } \hskip 0.5cm \includegraphics[width=0.39\textwidth]{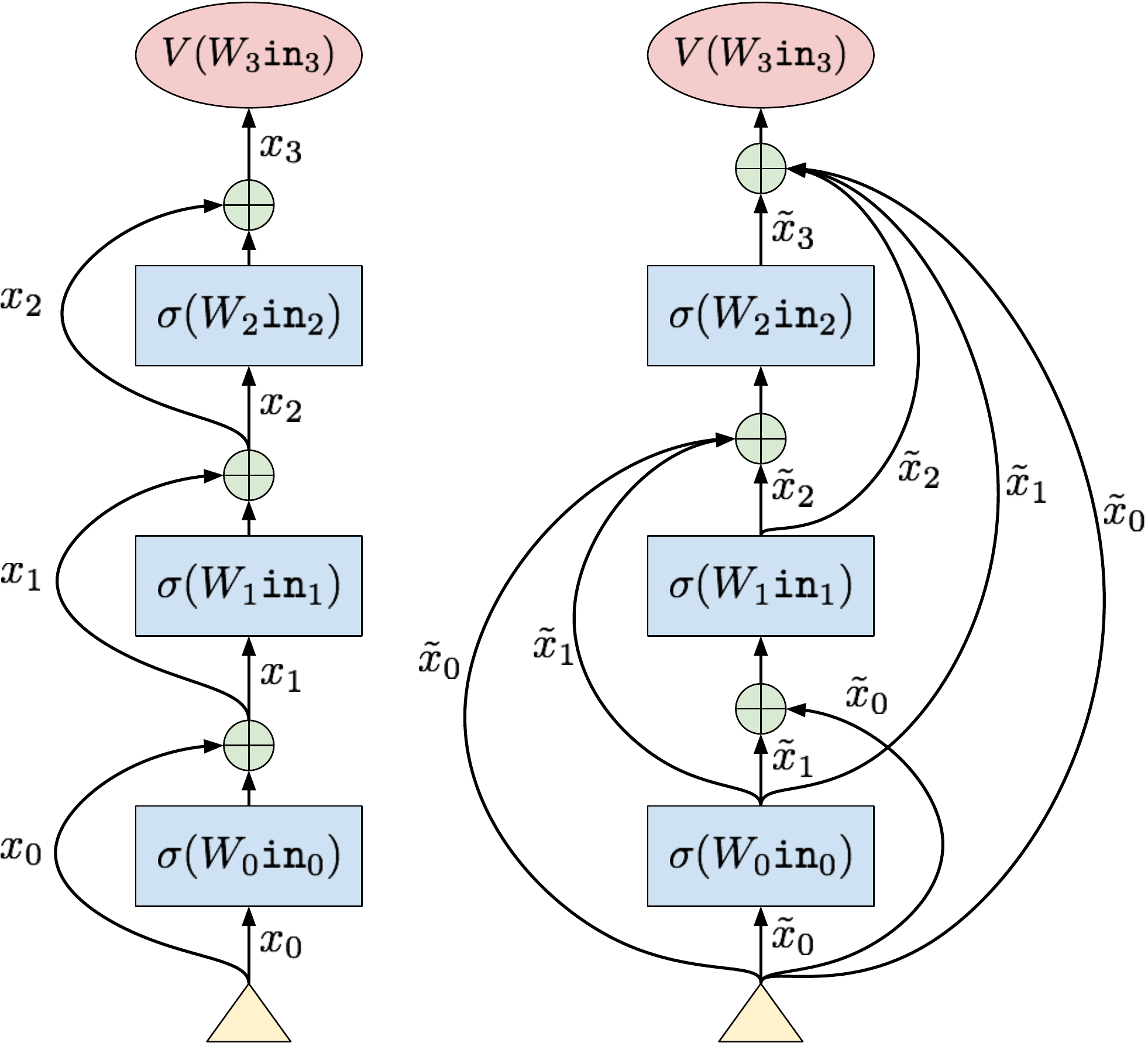}\\
$\ \ \ $ \hskip -0.9cm (a) \hskip 2.6cm (b) \hskip 0.22\textwidth (c) \hskip 2.0cm (d)
\caption{Left: the neurons and weights that are involved in the computations to update the red-dotted weight $w$ are highlighted in yellow. (a) Backpropagation; (b) Local Propagation -- the computations required to update the variables $x,\lambda$ (associated to the red neuron) are also considered. Right: (c) ResNet in the case of $H=3$, and (d) after the change of variables ($x_{\ell} \rightarrow \tilde{x}_{\ell}$) described in Sec.~\ref{sec:popular}. Greenish circles are sums, and the notation $\texttt{in}_{\ell}$ inside a rectangular block indicates the block input.}
\label{local}
\end{figure*}

One of the key features of LP is the locality in the gradient computations. Before going into further details, we remark that, in the context of this paper, the term \textit{locality} refers to those gradient computations with respect to a certain variable of layer $\ell$ that only involve units belonging (at most) to neighbouring layers. 
This is largely different from the usual case of the BackPropagation (BP) algorithm, where the gradient of the cost function with respect to a certain weight in layer $\ell$ is computed only after a forward and a backward steps that involve all the neural units of all layers (see Fig.~\ref{local} (a))
Differently, in the case of LP we have
\begin{eqnarray}
\label{w}\hskip-5mm \frac{\partial \mathcal{L}}{\partial W_{\ell}} &\hskip-3mm=\hskip-3mm& -\sum_{i=1}^{N} \left(\lambda_{\ell+1,i} \odot  \mathcal{G}'_{\ell+1,i} \odot \sigma'(W_{\ell}x_{\ell,i}) \right) x^{T}_{\ell,i}  \\
\label{uff}\hskip-5mm \frac{\partial \mathcal{L}}{\partial x_{\ell,i}} &\hskip-3mm=\hskip-3mm& \lambda_{\ell,i}\odot \mathcal{G}'_{\ell,i} \hskip-0.5mm - \hskip-0.5mm W_{\ell}^T \hskip -1mm \left(  \lambda_{\ell+1,i}\odot \hskip -0.5mm \mathcal{G}'_{\ell+1,i}  \odot \hskip -0.5mm \sigma'(W_{\ell}x_{\ell,i}) \right) \\
\label{x}\hskip-5mm \frac{\partial \mathcal{L}}{\partial x_{H,i}} &\hskip-3mm=\hskip-3mm& \lambda_{H,i}\odot \mathcal{G}'_{H,i} +  W_{H}^T \left( V'(W_{H}x_{H,i},y_i) \right)\\
\label{l}\hskip-5mm \frac{\partial \mathcal{L}}{\partial \lambda_{\ell,i}} &\hskip-3mm=\hskip-3mm& \mathcal{G}_{\ell,i}
\end{eqnarray}
where $\mathcal{G}'$, $\sigma'$ and $V'$ are the first derivatives of the respective functions, and $\odot$ denotes the Hadamard product. The  equations above  hold for all $i\in[1,n]$ and $\ell\in[1,H]$, with the exception of Eq. (\ref{uff}) that holds for $\ell\in[1,H-1]$.
It is evident that each partial derivative with respect to a variable associated to layer $\ell$ only involves terms that belong to the same layer (e.g., $\partial \mathcal{L} / \partial \lambda_{\ell,i}$) and also to either layer $\ell-1$ (as in $\partial \mathcal{L} / \partial W_{\ell}$) or layer $\ell+1$ (the case of $\partial \mathcal{L} / \partial x_{\ell,i}$), that is, gradient computations are \textit{local} (see Fig.~\ref{local} (b)).

This analysis reveals the full local structure of the algorithm for the discovery of saddle points.
The role of the local updates is twofold: first, they project the variables onto the feasible region defined by the $\mathcal{G}_{\ell,i}$ constraints; second, they allow the information attached to the supervised pairs to flow from the loss function $V(\cdot,\cdot)$ through the network. The latter consideration is critical, since the information can flow through a large number of paths, and many iterations could be required to keep the model projected onto the feasible region and efficiently learn the network weights. In Section~\ref{sec:exp} we will also explore the possibility of enforcing a $L_1$-norm regularizer (weighted by $\alpha > 0$) on each $x_{\ell,i}$, in order to help the model to focus on a smaller number of paths from input to output units, reducing the search space.

\paragraph{Parallel Computations over Layers.}
It is well known that in Backpropagation we have to perform a set of \textit{sequential} computations over layers to complete the forward stage, and only afterwards we can start to \textit{sequentially} compute the gradients, moving from the top layer down to the currently considered one (backward computations). 
Modern hardware (GPUs) can benefit by the parallelization of the matrix operations within each layer, 
while in the case of LP, the locality in the gradient computation allows us to go beyond that. We can promptly see from Eq.~(\ref{w}-\ref{l}) that we can trivially distribute all the computations associated to each layer $\ell$ in a different computational unit. Of course, the $\ell$-th computational unit needs to share the memory where some variables are stored with the $(\ell+1)$-th and $(\ell-1)$-th units (see Eq.~(\ref{w}-\ref{l})).

\paragraph{Deep Learning in the Adjoint Space.}
Learning in the space to which the variables $\mathcal{W}, \mathcal{X}, \Lambda$ belong introduces a particular information flow through the network. If, during the optimization stage, the architectural constraints of Eq.~(\ref{con}) are strongly violated, then the updates applied to the network weights are not related to the ground truths that are attached to the loss function $V(\cdot)$, and we can imagine that the gradients are just noise. Differently, when the constraints are fulfilled, the information traverses the network in a similar way to what happens in Backpropagation, i.e., in a noise-free manner. When the optimization proceeds, we progressively get closer to the fulfilment of the constraints, so that the noisy information is reduced. It is the learning algorithm itself that decides how to reduce the noise, in conjunction with the reduction of the loss on the supervised pairs. It has been shown that introducing a progressively reduced noise contribution to the gradient helps the Backpropagation algorithm to improve the quality of the solution, allowing very deep networks to be trained also when selecting low quality initialization of the weights \cite{neelakantan2015adding}. The LP natively embeds this property so that, differently from \cite{neelakantan2015adding}, the noise reduction scheme is not a hand-designed procedure. Moreover, the local gradient computations of LP naturally offer a setting that is more robust to the problem of vanishing gradients, which afflicts Backpropagation when training deep neural networks.

\paragraph{Recovering Backpropagation.}
The connections between the LP algorithm and Backpropagation become evident when imposing the stationary condition on the Lagrangian $\partial \mathcal{L} / \partial \lambda_{\ell,i}=0$ and $\partial \mathcal{L} / \partial x_{\ell,i} = 0$. For the purpose of this description, let $\mathcal{G}(\cdot)$ be the identity function. From Eq.~(\ref{l}), we can immediately see that the stationary condition $\partial \mathcal{L} / \partial \lambda_{\ell,i} = 0$ leads to the classic expression to compute the outputs of the neural units, $x_{\ell,i} = \sigma(W_{\ell-1}x_{\ell-1,i})$, that is associated to the forward step of Backpropagation. Differently, when imposing $\partial \mathcal{L} / \partial \lambda_{\ell,i}=0$ and defining $\delta_{\ell,i}=\lambda_{\ell,i} \odot \sigma'(W_{\ell-1}x_{\ell-1,i})$, Eq.~(\ref{w}) and Eq.~(\ref{uff}) can be respectively rewritten as
\begin{align*}
\frac{\partial \mathcal{L}}{\partial W_{\ell}} & = -\sum_{i=1}^N \delta_{\ell,i} \cdot x_{\ell-1,i}^T,\\ \delta_{\ell,i} & =\sigma'(W_{\ell-1}x_{\ell-1,i}) \odot \left( W_{\ell}^{T} \delta_{\ell+1,i} \right),
\end{align*}
that are the popular equations for updating weights and the Backpropagation deltas. 

From this perspective, the Backpropagation algorithm represents the optimum w.r.t. the stationary conditions connected to the $\lambda_{\ell,i}$ and the $x_{\ell,i}$ when compared with Local Propagation. However, by strictly searching only on the hyperplane where the Lagrangian is stationary w.r.t $\lambda_{\ell,i}$ and $x_{\ell,i}$, Backpropagation loses the locality and parallelization properties characterizing our algorithm since the gradients cannot rely anymore on the variables of neighboring layers only but depend on all the variables of the architecture.

\subsection{Epsilon-insensitive Constraints}
\label{sec:eps}
In order to facilitate the convergence of the optimization algorithm or to improve its numerical robustness, we can select different classes of $\mathcal{G}(\cdot)$ functions in Eq.~(\ref{con}). In this paper, we focus on the class of $\epsilon$-insensitive functions, and, in particular, on the following two cases $\mathcal{G} \in \{ \eps, \lineps \}$, 
\begin{eqnarray*}
\eps(a) &=& \max(|a| - \epsilon, 0) \\
\lineps(a) &=& \max(a, \epsilon) - \max(-a, \epsilon) \ .
\end{eqnarray*}
Both the functions are continuous, they are zero in $[-\epsilon, \epsilon]$, and they are linear out of such interval. However, $\eps(\cdot)$ is always positive, while $\lineps(\cdot)$ is negative for arguments smaller than  $-\epsilon$. When plugged into Eq.~(\ref{con}), they allow the architectural constraints to tolerate a bounded mismatch in the values of $x_{\ell,i}$ and $\sigma(W_{\ell-1}x_{\ell,i})$ ($\epsilon$-insensitive constraints). Let us consider two different examples indexed by $i$ and $j$, for which we get two similar valus $\sigma(W_{\ell-1}x_{\ell,i})$ and $\sigma(W_{\ell-1}x_{\ell,j})$ in a certain layer $\ell$. Then, for small values of $\epsilon$, the same value $x_{\ell,i}=x_{\ell,j}$ can be selected by the optimization algorithm, thus propagating the same signal to the units of the layer above. In other words, $\epsilon$-insensitive constraints introduce a simple form of regularization when training the network, that allows the network itself to not be influenced by small changes in the neuron inputs, thus stabilizing the training step. Notice that, at test stage, if we compute the values of $x_{\ell,i}$'s with the classic forward procedure, then the network will not take into account the $\mathcal{G}(\cdot)$ function anymore. If $\epsilon$ is too large, there will be a large discrepancy between the setting in which the weights are learned and the one in which they are used to make new predictions. This could end up in a loss of performances, but it is in line with what happens in the case of the popular Dropout \cite{srivastava2014dropout} when the selected drop-unit factor is too large.

A key difference between $\eps(\cdot)$ and $\lineps(\cdot)$ is the effect they have in the development of the Lagrange multipliers. It is trivial to see that, since $\eps(\cdot)$ is always positive, the multipliers $\lambda_{\ell,i}$ can only increase during the optimization (Eq.~(\ref{l})). In the case of $\lineps(\cdot)$, the multipliers can both increase or decrease. We found that $\eps(\cdot)$ leads to a more stable learning, where the violations of the constraints change more smoothly that in the case of $\lineps(\cdot)$. As suggested in \cite{platt} and as it is also popular in the optimization literature \cite{bertsekas2014constrained}, a way to improve the numerical stability of the algorithm is to introduce the so called Augmented Lagrangian, where $\mathcal{L}$ of Eq.~(\ref{lagrangian}) is augmented with an additive term $\rho \| \mathcal{G}_{\ell,i} \|^2$, for all $i,\ell$.

\section{Popular Neural Units}
\label{sec:popular}
The described constraint-based formulation of neural networks and the LP algorithm can be easily applied to the most popular neural units, thus offering a generic framework for learning in neural networks. It is trivial to rewrite Eq.~(\ref{con}) to model convolutional units and implement Convolutional Neural Networks (CNNs) \cite{lecun1998gradient}, and also the pooling layers can be straightforwardly described with constraints. We study in detail the cases of Recurrent Neural Networks (RNNs) and of Residual Networks (ResNets). In order to simplify the following descriptions, we consider the case in which we have only $N=1$ supervised pairs, and we drop the index $i$ to make the notation simpler.

\paragraph{Recurrent Neural Networks.}
At a first glance, RNNs \cite{hochreiter2001gradient} might sound more complicated to implement in the proposed framework. As a matter of fact, when dealing with RNNs and Backpropagation, we have to take care of the temporal unfolding of the network itself (Backpropagation Through Time)\footnote{What we study here can be further extended to the case of Long Short-Term Memories (LSTMs) \cite{hochreiter1997long}}. However, we can directly write the recurrence by means of architectural constraints and, we get that, for all time steps $t$ and for all layers $\ell \in [1,H]$,
\begin{equation*}
    \mathcal{G}\left(x^t_{{\ell}} - \sigma(W_{{\ell-1}} x^t_{{\ell-1}} + U_{{\ell-1}} x^{t-1}_{{\ell}})\right) = 0 \ ,
\end{equation*}
where $U_{\ell}$ is the matrix of the weights that tune the contribution of the state at the previous time step. The constraint-based formulation only requires to introduce constraints over all considered time instants. This implies that also the variables $x_{\ell}$ and the multipliers $\lambda_{\ell}$ are replicated over time (superscript $t$). The optimization algorithm has no differences with respect to what we described so far, and all the aforementioned properties of LP (Sec.~\ref{lp}) still hold also in the case of RNNs. 
While it is very well known that recurrent neural networks can deal only with sequential or DAG inputs structures (i.e. no cycles), LP architectural constraints show no ordering, since we ask for the overall fulfillment of the constraints. This property opens the door to the potential application of the proposed algorithm to problems dealing with generic graphical inputs, which is a very hot topic in the deep learning community \cite{scarselli2009graph,henaff2015deep,li2018learning}

\paragraph{Residual Networks.}
ResNets \cite{he2016deep} consist of several stacked residual units, that have been popularized by
their property of being robust with respect to the vanishing gradient problem, showing state-of-the art results in Deep Convolutional Neural Nets \cite{he2016deep} (without being limited to such networks).
The most generic form of a single residual unit is described in \cite{he2016identity},
\begin{equation}
x_{\ell} = z( h(x_{\ell-1}) + f(W_{\ell-1}x_{\ell-1})) \ .
\label{res}
\end{equation}
In the popular paper of \cite{he2016deep}, we have that $z(\cdot)$ is a rectifier (ReLu) and $h(\cdot)$ is the identity function, while $f(\cdot)$ is a non-linear function. On one hand, it is trivial to implement a residual unit as a constraint of LP once we introduce the constraint
\begin{equation}
\mathcal{G}(x_{\ell} - z( h(x_{\ell-1}) + f(W_{\ell-1}x_{\ell-1}))) = 0 \ .
\label{reslp}
\end{equation}
However, we are left with the question whether these units still provide the same advantages that 
they show in the case of backprop-optimized networks.
%
In order to investigate the ResNet properties, we focus on the identity mapping of \cite{he2016identity}, where $z(\cdot)$ and $h(\cdot)$ are both identity functions, and, for the sake of simplicity, $f(\cdot)$ is a plain neural unit with activation function $\sigma$,
\begin{equation}
x_{\ell} = x_{\ell-1} + \sigma(W_{\ell-1}x_{\ell-1}) \ ,
\label{ress}
\end{equation}
as sketched in Fig. \ref{local} (c).
This implementation of residual units is the one where it is easier to appreciate how the signal propagates through the network, both in the forward and backward steps. The authors of \cite{he2016deep} show that the signal propagates from layer $\ell$ to layer $L>\ell$ by means of additive operations, $x_{L}=x_{\ell} + \sum_{j=\ell}^{L-1}\sigma(W_{j}x_{j})$, while in common feedforward nets we have a set of products. Such property implies that the gradient of the loss function $V(W_{H}x_{H}, y)$ with respect to $x_{\ell}$ is
\begin{equation}
\frac{\partial{V}}{\partial{x_{\ell}}} = \frac{\partial{V}}{\partial x_{H}} \biggl( 1 + \frac{\partial}{\partial x_{\ell}} \sum_{j=\ell}^{H-1} \sigma(W_{j}x_{j})\biggr) \ ,
\label{vb}
\end{equation}
that clearly shows that there is a direct gradient propagation from $V$ to layer $\ell$ (due to the additive term 1). 
Due to the locality of the LP approach, this property is lost when computing the gradients of each architectural constraint, that in the case of the residual units of Eq. (\ref{ress}) are
\begin{equation}
\mathcal{G}(x_{\ell} - x_{\ell-1} - \sigma(W_{\ell-1} x_{\ell-1})) = 0 \ .
\label{resx}
\end{equation}
As a matter of fact, the loss $V(W_{H}x_{H}, y)$ will only have a role in the gradient with respect to variables $x_{H}$ and $W_{H}$, and no immediate effect in the gradient computations related to the other constraints/variables.
However, we can rewrite Eq. (\ref{resx}) by introducing $\tilde{x}_{\ell} = x_{\ell} - x_{\ell-1}$, that leads to $x_{\ell} = \tilde{x}_{\ell} + x_{\ell-1}$. 
By repeating the substitutions, we get $x_{\ell} = \sum_{j=0}^{\ell}\tilde{x}_{j}$, and Eq. (\ref{resx}) becomes\footnote{We set $\tilde{x}_{0} = x_{0}$ ($\tilde{x}_0$ is not a variable of the learning problem).}
\begin{equation}
\mathcal{G}\biggl(\tilde{x}_{\ell} - \sigma\Bigl(W_{\ell-1} \cdot \sum_{j=0}^{\ell-1}\tilde{x}_{j}\Bigr)\biggr) = 0\ ,
\label{resx2}
\end{equation}
where the arguments of the loss function change to $V(W_{H} \cdot \sum_{j=0}^{H} \tilde{x}_{j}, y)$.
Interestingly, this corresponds to a feed-forward network with activations that depend on the sum of the outputs of all the layers below, as shown in Fig. \ref{local} (d).
Given this new form of the first argument of $V(\cdot, \cdot)$, it is now evident that even if the gradient computations are local, the outputs of all the layers directly participate to such computations, formally
\begin{equation}
\frac{\partial{\mathcal{L}}}{\partial{\tilde{x}_{\ell}}} = \frac{\partial{V}}{\partial \tilde{x}_{\ell}} + \frac{\partial}{\partial \tilde{x}_{\ell}} \sum_{j=\ell}^{H}  \mathcal{G}_j  \ .
\end{equation}
Differently from Eq. (\ref{vb}), $\frac{\partial{V}}{\partial \tilde{x}_{\ell}}$ (that is the same $\forall \ell$) does not scale the gradients of the summation, so that the gradients coming from the constraints of all the hidden layers above $\ell$ are directly accumulated by sum.

\section{Experiments}
\label{sec:exp}

We designed a batch of experiments aimed at validating our simple local optimization approach to constraint-based networks. Our goal is to show that the approach is feasible and that the learned networks have generalization skills that are in-line with BackPropagation, also when using multiple hidden layers. In other words, we show that the new properties provided by our algorithm (i.e. locality and parallelization) does not correspond to a loss in performance w.r.t. Backpropagation, even if the search space has been augmented with unit activations and Lagrange multipliers variables.

We performed experiments on 7 benchmarks from the UCI repository \cite{Dua:2017}, and on the MNIST data (Table~\ref{tab:datasets}). 
The MNIST is partitioned into the standard training, validation and test sets, while in the case of the UCI data we followed the experimental setup of \cite{NIPS2018_7620}, where the authors used the training and validation partitions of  \cite{fernandez2014we} to tune the model parameters, and 4-folds to compute the final accuracy (averaged over the 4 test splits)\footnote{In the case of the Adult data we have only 1 test split.}. 
\begin{table}
\footnotesize
\centering
\caption{Number of patterns, of input features and of output classes of the datasets exploited for benchmarking our algorithm.}
	\begin{tabular}{lccc}
	\toprule
		\textsc{Dataset}    & \ \textsc{Examples} & \textsc{Dimensions} & \textsc{Classes} \\ 
		\midrule
		Adult      & 48842      & 14         & 2         \\
		Ionosphere & 351        & 33         & 2         \\
		Letter     & 20000      & 16         & 26        \\
		Pima       & 768        & 8          & 2         \\
		Wine       & 179        & 13         & 3         \\ 
		Ozone     & 2536 	  & 72  			&2       \\  
		Dermatology    & 366 &     34       &   6       \\
		MNIST      & 70000      & 784        & 10        \\		
	\bottomrule
	\end{tabular}
	
	\label{tab:datasets}
\end{table}

We evaluated several combinations of the involved parameters, varying them in: 
$\epsilon \in \{0, 10^{-4}, 10^{-3}, 10^{-2}\}$, $\rho \in \{10^{-2}, 10\}$, $c \in \{0, 0.001\}$, dropout keep-rate (BP only) $\in \{0.7, 0.8, 0.9 \}$,  $\alpha \in \{0, 10^{-8}, 10^{-5},  10^{-4}, 10^{-3}, 10^{-1}\}$. We used the Adam optimizer (TensorFlow), where the learning rate $\eta_w$ for updating variables $\mathcal{W}$ is $\in \{10^{-4}, 10^{-3},10^{-2}\}$, and the learning rate $\eta_z$ for updating $\mathcal{X}, {\Lambda}$ is $\in \{0.1\cdot \eta_w, 10 \cdot \eta_w\}$. 
We used the same initialization of the weight matrices to BP and LP.

Since similar behaviours are shown by both sigmoid and  ReLU activations, we exploited in our experiments only the former, in order to reduce the hyper-parameter search space. 
We trained our models for thousands epochs, measuring the accuracy on the validation data (or, if not available, a held-out portion of the training set) to select the best $\mathcal{W}$.

\begin{table*}
    \centering
          \caption{Performances of the same architectures optimized with BP and LP. Left: $H=1$ hidden layer (100 units); right: $H=3$ hidden layers (30 units each). Largest average accuracies are in bold.}
	\begin{tabular}{lcc | cc}
		\toprule
		& \multicolumn{1}{c}{   BP ($H=1$)} & \multicolumn{1}{c}{LP ($H=1$) } & \multicolumn{1}{|c}{BP ($H=3$) } & \multicolumn{1}{c}{LP ($H=3$) } \\
		\midrule
		Adult      &   $84.66$  $\pm 0.00$ &   $\textbf{85.43}$ \ $\pm 0.00$ & $84.91$ \ $ \pm 0.00$ &   $\textbf{85.34}$  $\pm 0.00$\\
		Iono. &    $91.48$  $\pm 0.57$ &  $91.48$  $\pm 2.95$ & $92.61$  $ \pm 0.57$ &   $\textbf{94.60}$  $\pm 1.86$ \\
		Letter     &    $94.20$ $\pm 0.31$ &   $\textbf{94.94}$  $\pm 0.05$ &  $\textbf{92.27}$  $ \pm 0.19$ &  $90.42$  $\pm 0.78$ \\
		Pima       &   $76.17$  $\pm 1.62$ &   $\textbf{77.21}$  $\pm 2.79$ &  $\textbf{76.56}$ $ \pm 2.42$ &  $75.91$ $\pm 1.54$  \\
		Wine       &    $97.16$  $\pm 1.88$ &   $\textbf{98.86}$  $\pm 1.14$ & $97.73$  $ \pm 2.78$ &   $\textbf{98.86}$  $\pm 1.97$ \\
		Ozone &     $ 97.04 $  $ \pm 0.26$ &  $ \textbf{97.12} $   $ \pm 0.13$ &  $ \textbf{97.28} $   $ \pm 0.13$ &  $ 97.20 $   $ \pm 0.17$ \\
		Derma.  &   $ 95.60 $   $ \pm 1.74$ &  $ \textbf{96.70} $   $ \pm 1.74$ &  $ 97.53 $   $ \pm 1.20$ &  $ \textbf{98.63} $  $ \pm 0.48$ \\
		\bottomrule
	\end{tabular}
	\label{bplp}
\end{table*}
\begin{figure}
 \centering
    \includegraphics[width=0.4\textwidth]{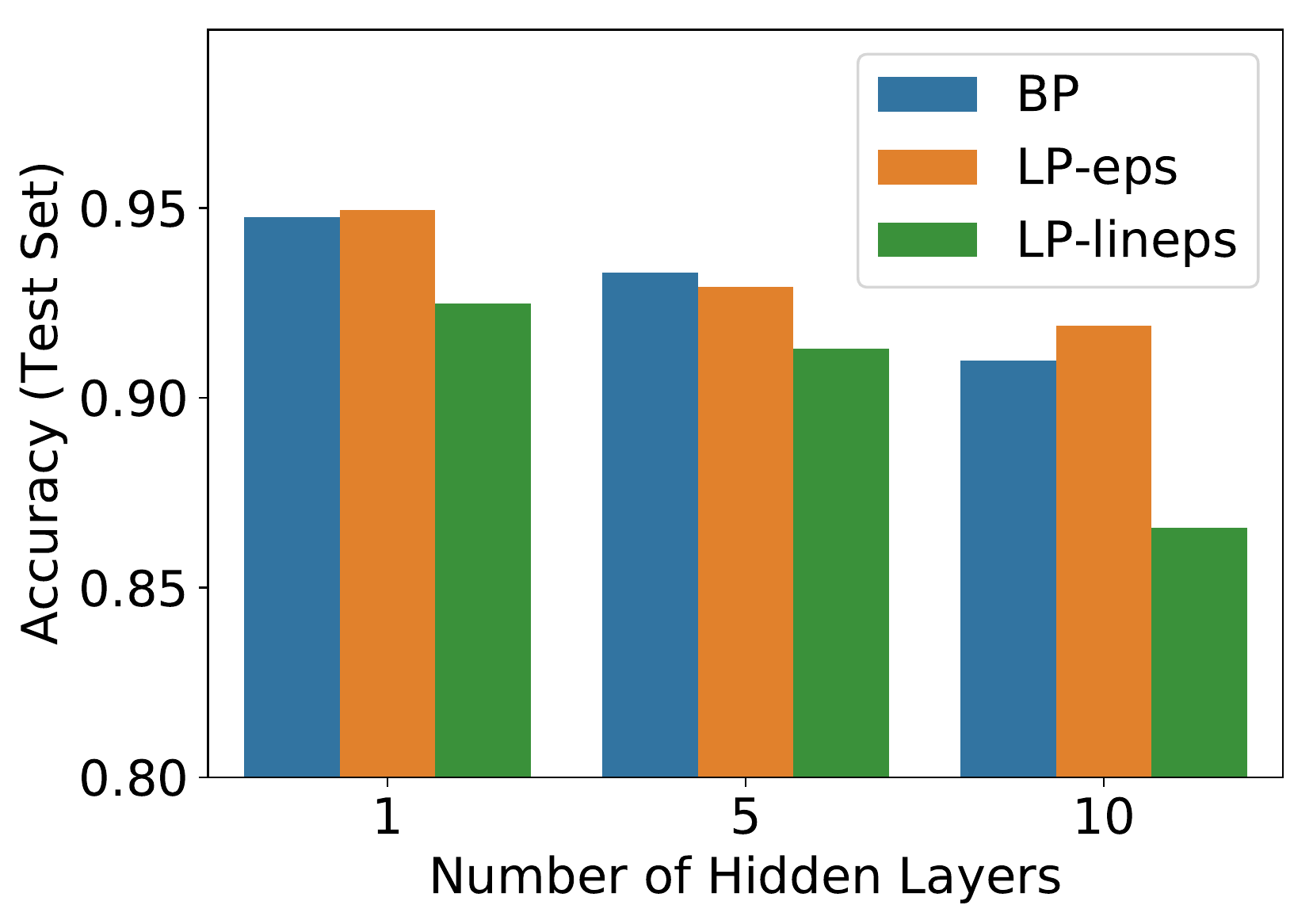}
    \caption{Accuracies of BP and LP (with different $\epsilon$-insensitive functions, $\eps$, $\lineps$) on the MNIST data.}
    \label{localf}
\end{figure}

We evaluated the accuracies of BP and LP focussing on the same pair of architectures (sigmoidal activation units), that is composed by a shallow net with 1 hidden layer of 100 units, and a deeper network with 3 hidden layers of 30 units each, reporting results in Table~\ref{bplp}. Both algorithms perform very similarly, with LP having some minor overall improvements over BP.

Similar conclusions can be drawn in the case of MNIST data, as shown in Fig.~\ref{localf}. In this case, we considered deeper networks with up to 10 hidden layers (10 neurons on each layer), and we also evaluated the impact of the different $\epsilon$-insensitive constraints of Section~\ref{sec:eps}. We considered 5 different runs, reporting the test accuracy corresponding to the largest result in the validation data. 
When using $\lineps(\cdot)$ as $\mathcal{G}(\cdot)$ function, we faced an oscillating behaviour of the learning procedure due to the inherent double-signed violation of the constraints. The Augmented Lagrangian ($\rho > 0$) resulted to be fundamental for the stability and for improving the convergence speed of LP.
\begin{figure}[!t]
	\centering
	\includegraphics[width=0.31\textwidth]{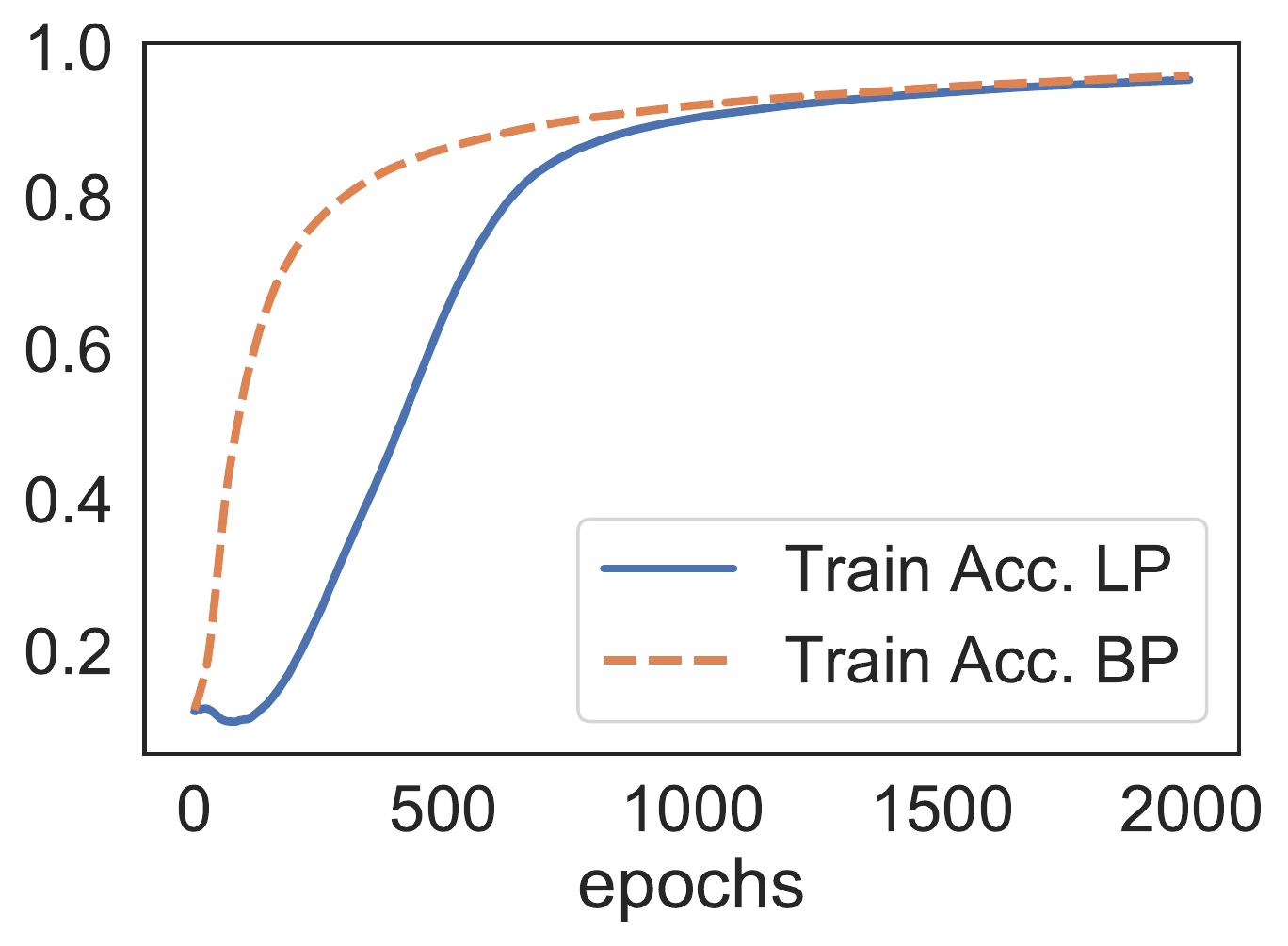}\hskip 2cm
	\includegraphics[width=0.31\textwidth]{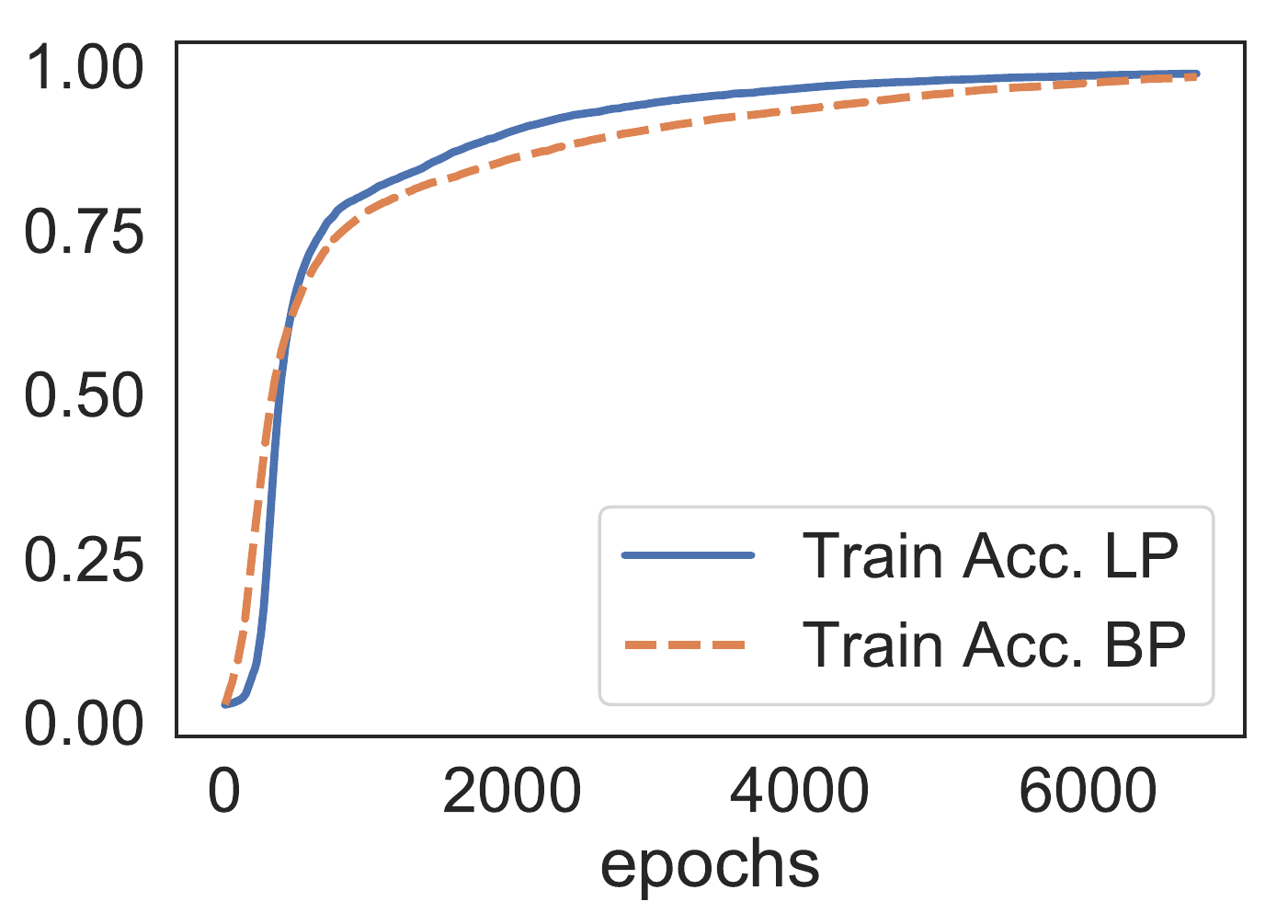}\\
	\caption{Convergence speed of BP and LP in the MNIST dataset (top), and in the Letter data (bottom).}
	\label{localxxx}
\end{figure}

Due to local nature of LP and to the larger number of variables involved in the optimization, we usually experimented an initial transitory stage in the optimization process, where the system is still far from fulfilling the available constraints, and the model accuracy is small, as shown in Fig.~\ref{localxxx} (a,b). This sometimes implies a larger number of iterations with respect to BackPropagation to converge to a solution (Fig.~\ref{localxxx} (a) - MNIST), as expected, but it is not always the case (Fig.~\ref{localxxx} (b) - Letter).

\begin{table*}[!ht]
    \centering
	\footnotesize
	\caption{Accuracies of LP when using ($\epsilon>0$) or not using ($\epsilon=0$) $\epsilon$-insensitive constraints (\textit{top-half}) and when using ($\alpha>0$) or not using ($\alpha=0$) $L_1$-norm-based regularization on the outputs of each layer (\textit{bottom-half}). We report the cases of the $\eps$ and $\lineps$ functions, and we compare architectures with $H=1$ hidden layer (100 units) and $H=3$ hidden layers (30 units each). 
	}
	\hskip -0.5cm
	\begin{tabular}{@{\hspace{0.0\tabcolsep}}p{0.6cm}p{1.5cm}@{\hspace{0.7\tabcolsep}}p{1.5cm}@{\hspace{0.7\tabcolsep}}|@{\hspace{0.7\tabcolsep}}p{1.7cm}@{\hspace{0.01\tabcolsep}}p{1.7cm}@{\hspace{0.1\tabcolsep}}||@{\hspace{0.7\tabcolsep}}p{1.5cm}@{\hspace{0.7\tabcolsep}}p{1.5cm}@{\hspace{0.7\tabcolsep}}|@{\hspace{0.7\tabcolsep}}p{1.7cm}@{\hspace{0.01\tabcolsep}}p{1.7cm}@{\hspace{0.0\tabcolsep}}}
		\toprule
		& &\multicolumn{2}{c}{$H = 1$}  & & & \multicolumn{2}{c}{$H = 3$}\\
		\midrule
		&     $ \epsilon = 0$ {\tiny ($\eps$)} &      $\epsilon > 0$  {\tiny ($\eps$)}&     $ \epsilon = 0$  {\tiny ($\lineps$)} &      $\epsilon > 0$  {\tiny ($\lineps$)} & $ \epsilon = 0$ {\tiny ($\eps$)} &      $\epsilon > 0$  {\tiny ($\eps$)}&     $ \epsilon = 0$  {\tiny ($\lineps$)} &      $\epsilon > 0$  {\tiny ($\lineps$)} \\
		\midrule
		Adult       &  $ 85.33 $ \hphantom{{\tiny  $ \pm 0.00$}} &  $ 85.33 $ \hphantom{{\tiny  $ \pm 0.00$}} &  $ \textbf{85.43} $ \hphantom{{\tiny  $ \pm 0.00$}} &  $ 85.27 $ \hphantom{{\tiny  $ \pm 0.00$}} &  $ \textbf{85.34} $ \hphantom{{\tiny  $ \pm 0.00$}} &  $ 85.25 $ \hphantom{{\tiny  $ \pm 0.00$}} &  $ \textbf{85.25} $ \hphantom{{\tiny  $ \pm 0.00$}} &  $ 85.23 $ \hphantom{{\tiny  $ \pm 0.00$}} \\

		Iono  &  $  \textbf{91.48} $ {\tiny  $ \pm 2.95$} &  $  91.19 $ {\tiny  $ \pm 2.71$} &  $ 90.34 $ {\tiny  $ \pm 0.98$} &  $  \textbf{91.19} $ {\tiny  $ \pm 2.18$} &  $  \textbf{94.60} $ {\tiny  $ \pm 1.86$} &  $ 94.32 $ {\tiny  $ \pm 1.14$} &  $  \textbf{92.61} $ {\tiny  $ \pm 1.27$} &  $ 91.19 $ {\tiny  $ \pm 0.94$} \\
		Letter     &  $  \textbf{94.94} $ {\tiny  $ \pm 0.05$} &  $ 94.85 $ {\tiny  $ \pm 0.06$} &  $ 93.71 $ {\tiny  $ \pm 0.20$} &  $ \textbf{94.00} $ {\tiny  $ \pm 0.27$} &  $  \textbf{87.60} $ {\tiny  $ \pm 0.48$} &  $ 87.54 $ {\tiny  $ \pm 0.48$} &  $  \textbf{90.42} $ {\tiny  $ \pm 0.78$} &  $ 90.39 $ {\tiny  $ \pm 0.27$} \\
		Pima       &  $ 75.39 $ {\tiny  $ \pm 2.03$} &  $  \textbf{77.21} $ {\tiny  $ \pm 2.79$} &  $ 75.00 $ {\tiny  $ \pm 1.77$} &  $  \textbf{75.78} $ {\tiny  $ \pm 1.97$} &  $ 75.52 $ {\tiny  $ \pm 1.91$} &  $  \textbf{75.91} $ {\tiny  $ \pm 3.34$} &  $ 74.48 $ {\tiny  $ \pm 0.90$} &  $  \textbf{75.91} $ {\tiny  $ \pm 1.54$} \\
		Wine      &  $ 97.73 $ {\tiny  $ \pm 1.61$} &  $  \textbf{98.86} $ {\tiny  $ \pm 1.14$} &  $ 98.30 $ {\tiny  $ \pm 1.88$} &  $ 98.30 $ {\tiny  $ \pm 1.88$} &  $ 97.73 $ {\tiny  $ \pm 3.94$} &  $ 97.73 $ {\tiny  $ \pm 2.78$} &  $  \textbf{98.86} $ {\tiny  $ \pm 1.97$} &  $ 97.73 $ {\tiny  $ \pm 1.61$} \\
		Ozone    &  $97.04 $ {\tiny  $ \pm 0.13$} &  $ 97.04 $ {\tiny  $ \pm 0.13$} &  $ 96.96 $ {\tiny  $ \pm 0.30$} &  $ \textbf{97.12} $ {\tiny  $ \pm 0.13$} &  $ 97.08 $ {\tiny  $ \pm 0.14$} &  $  \textbf{97.20} $ {\tiny  $ \pm 0.17$} &  $  \textbf{97.16} $ {\tiny  $ \pm 0.22$} &  $ 97.04 $ {\tiny  $ \pm 0.34$} \\ 
		Derma. &  $  \textbf{95.60} $ {\tiny  $ \pm 0.78$} &  $ 95.33 $ {\tiny  $ \pm 2.11$} &  $ 95.60 $ {\tiny  $ \pm 1.74$} &  $  \textbf{96.70} $ {\tiny  $ \pm 1.74$} &  $ \textbf{98.63} $ {\tiny  $ \pm 0.48$} &  $ 97.80 $ {\tiny  $ \pm 0.78$} &  $ 97.80 $ {\tiny  $ \pm 0.78$} &  $  \textbf{98.08} $ {\tiny  $ \pm 0.91$} \\

		\midrule
		&     $ \alpha = 0$ {\tiny ($\eps$)} &      $\alpha > 0$  {\tiny ($\eps$)}&     $ \alpha = 0$  {\tiny ($\lineps$)} &      $\alpha > 0$  {\tiny ($\lineps$)} & $ \alpha = 0$ {\tiny ($\eps$)} &      $\alpha > 0$  {\tiny ($\eps$)}&     $ \alpha = 0$  {\tiny ($\lineps$)} &      $\alpha > 0$  {\tiny ($\lineps$)} \\
		\midrule
		Adult      &  $ 85.33 $ \hphantom{{\tiny  $ \pm 0.00$}} &  $ 85.33 $ \hphantom{{\tiny  $ \pm 0.00$}} &  $ 85.27 $ \hphantom{{\tiny  $ \pm 0.00$}} &  $  \textbf{85.43} $ \hphantom{{\tiny  $ \pm 0.00$}} &  $  \textbf{85.34} $ \hphantom{{\tiny  $ \pm 0.00$}}&  $ 84.93 $ \hphantom{{\tiny  $ \pm 0.00$}}&  $ 85.23 $ \hphantom{{\tiny  $ \pm 0.00$}} &  $  \textbf{85.25} $ \hphantom{{\tiny  $ \pm 0.00$}} \\
		Iono  &  $ 90.63 $ {\tiny  $ \pm 0.49$} &  $ \textbf{91.48} $ {\tiny  $ \pm 2.95$} &  $  \textbf{91.19} $ {\tiny  $ \pm 2.18$} &  $ 90.06 $ {\tiny  $ \pm 2.71$} &  $ 90.34 $ {\tiny  $ \pm 1.70$} &  $ \textbf{94.60} $ {\tiny  $ \pm 1.86$} &  $ 89.77 $ {\tiny  $ \pm 2.27$} &  $  \textbf{92.61} $ {\tiny  $ \pm 1.27$} \\
		Letter       &  $ 94.85 $ {\tiny  $ \pm 0.27$} &  $  \textbf{94.94} $ {\tiny  $ \pm 0.05$} &  $ 93.78 $ {\tiny  $ \pm 0.38$} &  $  \textbf{94.00} $ {\tiny  $ \pm 0.27$} &  $  \textbf{87.60} $ {\tiny  $ \pm 0.48$} &  $ 87.54 $ {\tiny  $ \pm 0.48$} &  $  \textbf{90.42} $ {\tiny  $ \pm 0.78$} &  $ 88.88 $ {\tiny  $ \pm 0.45$} \\
		Pima       &  $ 75.39 $ {\tiny  $ \pm 2.03$} &  $  \textbf{77.21} $ {\tiny  $ \pm 2.79$} &  $  \textbf{75.78} $ {\tiny  $ \pm 1.97$} &  $ 75.00 $ {\tiny  $ \pm 1.77$} &  $  \textbf{75.91} $ {\tiny  $ \pm 3.34$} &  $ 75.65 $ {\tiny  $ \pm 3.62$} &  $  \textbf{75.91} $ {\tiny  $ \pm 1.54$} &  $ 75.65 $ {\tiny  $ \pm 1.89$} \\
		Wine      &  $ 98.86 $ {\tiny  $ \pm 1.97$} &  $ 98.86 $ {\tiny  $ \pm 1.14$} &  $  \textbf{98.30} $ {\tiny  $ \pm 1.88$} &  $  97.73 $ {\tiny  $ \pm 1.61$} &  $  \textbf{97.73} $ {\tiny  $ \pm 2.78$} &  $ 96.59 $ {\tiny  $ \pm 3.77$} &  $  \textbf{98.86} $ {\tiny  $ \pm 1.97$} &  $ 97.73 $ {\tiny  $ \pm 1.61$} \\
		Ozone    &  $ 97.00 $ {\tiny  $ \pm 0.11$} &  $  \textbf{97.04} $ {\tiny  $ \pm 0.13$} &  $  \textbf{97.12} $ {\tiny  $ \pm 0.13$} &  $ 96.96 $ {\tiny  $ \pm 0.30$} &  $ 97.08 $ {\tiny  $ \pm 0.14$} &  $  \textbf{97.20} $ {\tiny  $ \pm 0.17$} &  $ 97.08 $ {\tiny  $ \pm 0.14$} &  $  \textbf{97.16} $ {\tiny  $ \pm 0.22$} \\
		Derma.  & $ 95.60 $ {\tiny  $ \pm 1.10$} &  $ 95.60 $ {\tiny  $ \pm 0.78$} &  $ 95.88 $ {\tiny  $ \pm 1.43$} &  $  \textbf{96.70} $ {\tiny  $ \pm 1.74$} &  $ 97.80 $ {\tiny  $ \pm 0.78$} &  $  \textbf{98.63} $ {\tiny  $ \pm 0.48$} &  $  \textbf{98.08} $ {\tiny  $ \pm 0.91$} &  $ 97.80 $ {\tiny  $ \pm 0.78$} \\
		\bottomrule
	\end{tabular}
	\label{eee}
\end{table*}

In order to better understand how LP behaves, we deeply explored the previously discussed results of Table~\ref{bplp}. First, we evaluated the role of the $\epsilon$-insensitive constraints, reporting in Table~\ref{eee} (top-half) differentiated results for the case in which $\epsilon=$ and $\epsilon>0$. Then, we explored the effect of including  the $L_1$-norm regularization term, as shown in Table~\ref{eee} (\textit{bottom-half}) ($\alpha=0$ means no-$L_1$-regularization).
In the case of shallow networks, $\epsilon>0$ offers performances that, on average, are preferable or on-par to the case in which no-tolerance is considered (both for $\eps$ and $\lineps$). This consideration is not evident in the case of deeper nets, where a too strong insensitivity might badly propagate the signal from the ground truth to the lower layers. We notice that while this is evident in the case of UCI data, we did not experienced this behaviour in the case of MNIST of the aforementioned Fig.~\ref{localf}, where the best accuracies where usually associated with $\epsilon>0$. This might be due to the smallest redundancy of information in the UCI data with respect to MNIST. When focussing on the effect of the $L_1$-norm-based regularization (Table~\ref{eee}, bottom-half), we can easily see that such regularization helps in several cases, suggesting that it is a useful feature that should be considered in validating LP-based networks. This is due to the sparsification effect that emphasizes only a few neurons per layer, allowing LP to focus on a smaller number of input-output paths.

\section{Conclusions and Future Work}
\label{sec:concl}
This paper presented an uncommn way of interpreting the architecture of neural networks and learning their parameters, based on the so-called architectural constraints. It has been shown that the Lagrangian formulation in the adjoint space leads to a fully local algorithm, LP, that naturally emerges when searching for saddle points. An experimental analysis on several benchmarks assessed the feasibility of the proposed approach, whose connections with popular neural models has been described.
Despite its simplicity and its strongly parallelizable computations, LP introduces additional variables to the learning problem. We are currently studying an online implementation that is expected to strongly reduce the number of involved variables. LP opens the road to further investigations on other neural architectures, such as the ones that operate on graphs.

\section*{ Acknowledgments}
This work was partly supported by the PRIN 2017 project RexLearn, funded by the Italian Ministry of Education, University and Research (grant no. 2017TWNMH2).

\bibliography{biblo}
\bibliographystyle{IEEEtran.bst}

\end{document}